\pdfoutput=1

\documentclass[11pt]{article}

\usepackage{acl}

\usepackage{times}
\usepackage{latexsym}

\usepackage[T1]{fontenc}

\usepackage[utf8]{inputenc}

\usepackage{microtype}
\usepackage{url,color}
\usepackage{amsmath}
\usepackage{verbatim}
\usepackage{booktabs}
\usepackage{bm}
\usepackage{graphicx}
\usepackage{natbib}
\usepackage{multirow,array}
\newcommand{\secref}[1]{Section \ref{#1}}
\newcommand{\figref}[1]{Figure \ref{#1}}

\newcommand{\tabref}[1]{Table \ref{#1}}

\title{Leaner and Faster: Two-Stage Model Compression for Lightweight Text-Image Retrieval}

\author{Siyu Ren \hspace*{1cm} Kenny Q. Zhu\textsuperscript{\rm}\thanks{\hspace{2mm}The 
corresponding author.}\\
	Shanghai Jiao Tong University\\
	Shanghai, China\\
	roy0702@sjtu.edu.cn, kzhu@cs.sjtu.edu.cn}

\begin{document}
\maketitle
\begin{abstract}
Current text-image approaches~(e.g., CLIP) typically adopt dual-encoder architecture 
using pre-trained vision-language representation. 
However, these models still pose non-trivial memory requirements and 
substantial incremental indexing time, which makes them less practical on
mobile devices. In this paper, we present an effective two-stage framework 
to compress large pre-trained dual-encoder for lightweight text-image retrieval. 
The resulting model is smaller~(39\% of the original), 
faster~(1.6x/2.9x for processing image/text respectively), yet performs on par with or better than the original full model on Flickr30K and MSCOCO benchmarks. 
We also open-source an accompanying realistic mobile image search application.\end{abstract}

\section{Introduction}
Text-image retrieval is the task aiming at retrieving a list of relevant images from 
a large set of images given a textual query specified by the user. 
Recently, large-scale vision-language pretraining~(VLP) has spawned models~\cite{tan-bansal-2019-lxmert,oscar,clip} that established state-of-the-art results in various vision-language tasks~\cite{vqa,nlvr}, including text-image retrieval. Existing VLP models for text-image retrieval can be divided into two categories: cross-encoder architecture and dual-encoder architecture. Cross-encoder models show better retrieval accuracy by allowing fine-grained cross-modal attention among image and text. However, they are prohibitively slow to apply to the entire image pool because each image has to go through the deep Transformer again whenever a new text query comes in. Moreover, most cross-encoder models rely on external object detection models~\cite{fasterrcnn} to extract visual features, which further increase memory consumption. On the other hand, dual-encoder models are more scalable in that they allow pre-computing image representations as 
reusable vectors independent of the text queries. These image vectors can be
indexed and efficiently retrieved at runtime using Approximate Nearest Neighbor~(ANN) 
search~\cite{faiss}. As long as the image pool remains unchanged, 
the image encoder is not required. 

\begin{figure*}[t!]
	\centering
	\scalebox{0.94}{\includegraphics[width=2.0\columnwidth]{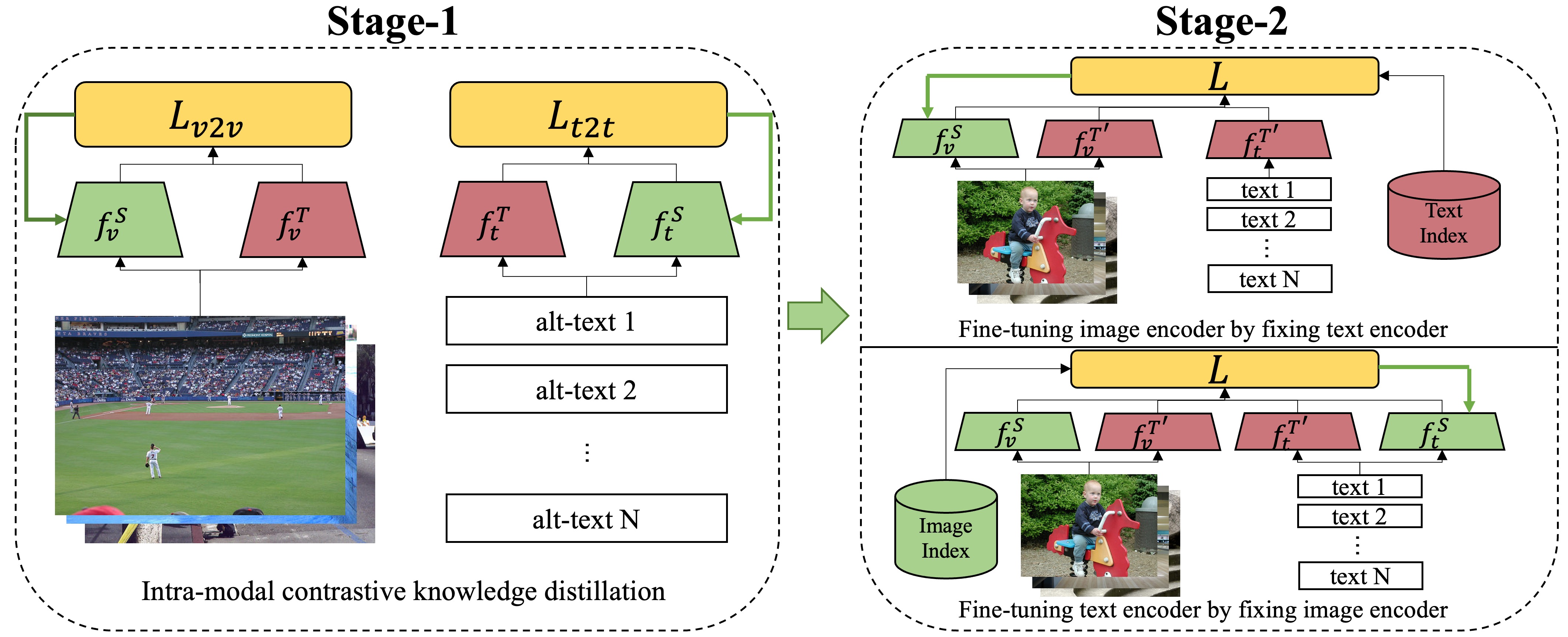}}
	\caption{In stage-1~(\secref{stage1}), we perform  \textit{intra-modal} contrastive knowledge distillation. In stage-2~(\secref{stage2}), we sequentially fine-tune  $f_v^{S}$ and $f_t^{S}$ with knowledge distillation~(KD) and corpus-level hard negative~(HN) mining via pre-computed index. The total loss $L$ is the sum of $L_{t2v}$, $L_{v2t}$, $L_{KD}$, and $L_{HN}$. The thin black arrows represent the input/output flows and 
the solid green arrows indicate the gradient flows.
	} \label{fig:overview}
\end{figure*}

However, a more practical scenario calls for dynamic indexing of new images into the pool~(e.g., private photo collections on mobile devices),
which requires both the image encoder and the text encoder to be resident in memory. 
This makes the above approach less practical on mobile devices with limited
memory and processing power. 
Unfortunately, little attention has been paid to fulfill this need. 
In this paper, we show that a large dual-encoder model can be compressed into 
a much smaller and faster counterpart while retaining its retrieval accuracy 
using a novel two-stage compression framework. 
In the first stage, we make use of abundant non-paired texts/images to separately 
compress text or image encoder with an effective intra-modal contrastive knowledge 
distillation scheme. In the second stage, we sequentially fine-tune the distilled image or
text encoder on paired text-image data with comprehensive learning objectives.
Using CLIP~\cite{clip} as the target model, our compressed models deliver comparable performance on MSCOCO and Flickr30K while being just 39\% of the original size 
and 1.6x/2.9x times faster for processing image/text. 
Detailed ablation study shows the effectiveness of each component in the 
compression framework and their synergistic effects. 

Our contributions are three-folds: 1)~an effective compression framework tailored for lightweight text-image retrieval; 2)~a leaner and faster model with competitive accuracy; 3)~open-sourced models and text-to-image search mobile applications on both iOS and Android at \url{https://github.com/DRSY/MoTIS}.

\section{Related Work}
\textbf{Cross-encoder.}~Cross-encoder architecture~\cite{tan-bansal-2019-lxmert,uniter,oscar} adopts a single Transformer network~\cite{NIPS2017_3f5ee243} which is able to process inputs from different modalities, e.g., images and texts. Benefitting from the self-attention mechanism, the hidden states of images and texts 
interact with each other at the patch/token-level, therefore yielding state-of-the-art retrieval accuracy. Though effective, these models suffer from huge memory consumption and inference latency, making them inpractical in time-sensitive real-world scenarios.

\textbf{Dual-encoder.}~In contrast to cross-encoder, dual-encoder architecture~\cite{clip,align} trains two seperate encoders for vision and language modalities. The exact choices of encoder architecture may be different. For example, CLIP utilizes Transformers for both visual and text encoders, while ALIGN~\cite{align} uses pre-trained BERT as text encoder and EfficientNet as visual encoder. In dual encoder, interactions between different 
modalities take place only at the final encoder layer, resulting in slightly worse performance compared to cross-encoders. Nevertheless, this late-interaction scheme of dual-encoder allows for efficient similarity computation, thus rendering it suitable for prividing real-time searching.

\section{Approach}


\subsection{Background on Dual-Encoder}
\label{pre}
Dual-encoder architecture employs two separate neural networks to encode inputs from different modalities and map them to a shared  space. 

We denote the image encoder as $f_v$ and the text encoder as $f_t$ in the context of text-image retrieval. To train $f_v$ and $f_t$, it is common to adopt an objective that pushes the embeddings of matched text-image pairs closer while pushing those of non-matched text-image pairs apart. Specifically, \textbf{C}ontrastive \textbf{L}anguage-\textbf{I}mage \textbf{P}retraining~(CLIP)~\cite{clip} optimizes an InfoNCE~\cite{infonce} loss:
\begin{align}
	\mathcal{L}_{t2v}=-\frac{1}{N}\sum_{i=1}^{N}\log{\frac{e^{f_t(x_i)^\top f_v(y_i)/\tau}}{\sum_{j=1}^{N}e^{f_t(x_i)^\top f_v(y_j)/\tau}}}
	\label{infonce}
\end{align}
Here, $f_t(x_i)$ and $f_v(y_j)$ are the L2-normalized embeddings of text in the $i$-th pair and image in the $j$-th pair. $N$ is the mini-batch size and $\tau$ is the temperature to scale the logits. The final objective is the sum of $L_{t2v}$ and its symmetric version $L_{v2t}$.

\subsection{Two-Stage Model Compression}
\label{sec2}
Despite good retrieval accuracy, models like CLIP still pose non-trivial memory 
footprint and inference time, which is undesirable for low-resource devices 
such as smart phones. 

To tackle this issue, we propose a two-stage compression framework to make large dual-encoder model smaller and faster while retaining its accuracy. A schematic overview is illustrated in \figref{fig:overview}. The first stage is \textit{task-agnostic}, where we leverage massively available non-paired texts/images to separately compress the text/image encoder using an intra-modal contrastive knowledge distillation scheme. The second stage is \textit{task-specific}, where we sequentially fine-tune the distilled image and text encoder using a combination of multiple techniques. We denote the image and text encoder of the large dual-encoder as $f_v^{T}$ and $f_t^{T}$ and those of the compressed model as $f_v^{S}$ and $f_t^{S}$.

\subsubsection{Stage-1}
\label{stage1}
The extremely large scale of text-image pairs~(e.g., 400 million used to train CLIP) makes it possible to make up for the noise in data and train over-parametrized large dual-encoder~(i.e., $f_v^{T}$ and $f_t^{T}$) from scratch to learn aligned visual and language representations. However, it is difficult to train small model~(i.e., $f_v^{S}$ and $f_t^{S}$) with lower capacity using the same \textit{inter-modal} learning scheme.

To circumvent this issue, we propose to exploit massively available non-paired data from the web and optimize an \textit{intra-modal} contrastive objective that aligns the output embeddings of $f^{S}$ and pretrained $f^{T}$, which can be seen as a form of knowledge distillation~\cite{hinton2015distilling}. Here we take visual modality as an example. Given a collection of images $\{y_i\}_{i=1}^N$, we feed them to both $f_v^{S}$ and $f_v^{T}$ to produce two sets of image embeddings $\{f_v^{S}(y_i)\}_{i=1}^{N}$ and $\{f_v^{T}(y_i)\}_{i=1}^{N}$. Then we optimize the following contrastive objective for updating $f_v^{S}$:
\begin{align}
	\mathcal{L}_{v2v}=-\frac{1}{N}\sum_{i=1}^{N}\log{\frac{e^{f_v^{S}(y_i)^\top f_v^{T}(y_i)/\tau}}{\sum_{j=1}^{N}e^{f_v^{S}(y_i)^\top f_v^{T}(y_j)/\tau}}}
\end{align}
The same formulation is symmetrically applied to language modality to obtain $L_{t2t}$ for updating $f_t^{S}$:
\begin{align}
	\mathcal{L}_{t2t}=-\frac{1}{N}\sum_{i=1}^{N}\log{\frac{e^{f_t^{S}(x_i)^\top f_t^{T}(x_i)/\tau}}{\sum_{j=1}^{N}e^{f_t^{S}(x_i)^\top f_t^{T}(x_j)/\tau}}}
\end{align}
Essentially, $f_v^{S}$/$f_t^{S}$ is trained to recover the representation power of $f_v^{T}$/$f_t^{T}$ in a decoupled manner.

\subsubsection{Stage-2}
\label{stage2}
After training $f_v^{S}$ and $f_t^{S}$ using general-domain data, 
it is necessary to adapt the learned representations to downstream tasks 
using in-domain data. First, we fine-tune $f_v^{T}$ and $f_t^{T}$ on paired 
text-image data $D=\{(x_i, y_i)\}_{i=1}^{N}$ using standard InfoNCE loss~(\secref{pre}). In the experiments, we found that jointly fine-tuning image 
and text encoder results in retrieval performance even worse than no fine-tuning at all. 
Therefore, we choose to sequentially fine-tune $f_v^{T}$/$f_t^{T}$ by fixing the other one. The resulting fine-tuned encoders are denoted as $f_v^{T^{\prime}}$ and $f_t^{T^{\prime}}$ and are henceforth kept fixed. Next, for training $f_v^{S}$ and $f_t^{S}$, we propose several techniques  essential to successful compression:

\noindent
\textbf{Knowledge Distillation~(KD).} In addition to the standard InfoNCE loss, we design two kinds of knowledge distillation objectives to learn from $f_v^{T^{\prime}}$ and $f_t^{T^{\prime}}$. One is the  Kullback-Leibler divergence between image-text matching distribution predicted by $f_v^{T^{\prime}}$ and $f_t^{T^{\prime}}$ and the one predicted by $f_v^{S}$ and $f_t^{S}$. This resembles previous response-based knowledge distillation~\cite{hinton2015distilling}. The other is the same contrastive objective  defined in \secref{stage1}. It indirectly encourages the alingment between visual and language representations.

\noindent
\textbf{Sequential Finetuning~(SF).} Similar to how we get $f_v^{T^{\prime}}$ and $f_t^{T^{\prime}}$, we also fine-tune $f_v^{S}$ and $f_t^{S}$ in a sequential manner. Concretely, we first let the compressed model share the same text encoder with the target dual-encoder and only fine-tune its image encoder. After that, we then fix the image encoder and fine-tune its text encoder in the same way.

\noindent
\textbf{Hard Negative Mining~(HN).} Prior works on contrastive representation learning~\cite{simclr,simcse}  typically exploit in-batch negative samples. Though efficient, image-text pairs in a batch are randomly sampled and are likely to be trivially unrelated. Models trained in such a way may fail in cases where candidates are similar. To achieve more accurate retrieval, we mine hard negatives from the entire corpus. In our sequential fine-tuning setting, we first use $f_t^{T^{\prime}}$ to compute embeddings of all texts in the corpus and index them with Faiss~\cite{faiss}. During training $f_v^{S}$, for each image $y_i$ we use $f_v^{S}(y_i)$ as query to the index and obtain its top-k texts as negative samples. Afterward, we use the trained $f_v^{S}$ to compute embeddings of all images in the corpus and build the index. During training $f_t^{S}$, for each text $x_i$ we use $f_t^{S}(x_i)$ as query to the index and get its top-k images as negative samples.

The complete training objective of stage-2 is  defined as $\mathcal{L}=\mathcal{L}_{t2v}+\mathcal{L}_{v2t}+\mathcal{L}_{KD}+\mathcal{L}_{HN}$.

\section{Experiment}

\begin{table*}[t!]
	
	\centering
	
	\begin{tabular}{cc|ccccccccc}
		
		\toprule
		
		\multirow{2}{*}{Image}&\multirow{2}{*}{Text} & \multicolumn{3}{c}{MSCOCO~(1K)} &\multicolumn{3}{c}{MSCOCO~(5K)}  &\multicolumn{3}{c}{Flickr 30K}  \\
		
		&  &R@1 &R@5  &R@10  &R@1  &R@5  &R@10 &R@1  &R@5  &R@10 \\
		
		\cline{1-11}

		$f_v^{T}$ & $f_t^{T}$ &46.9  &77.3  &87.3   &28.0  &52.9  &64.5  & 55.2 &80.3  &87.8 \\
		
		$f_v^{T^\prime}$ & $f_t^{T\prime}$ &61.0 &87.9 &94.7 &40.9 &67.6  &77.9 &58.0  &82.3 &89.1  \\
		
		\cline{1-11}
				$f_v^{S}$ & $f_t^{\text{TinyBERT}}$ &41.4 &76.7 &88.1 &21.3 &47.2 &61.0  &30.2  &59.1  &71.2 \\
				$f_v^{S}$ & $f_t^{S_4}$ &62.0&88.0 &94.4 &42.0 &69.2 &79.0  &55.0  &81.3  &88.4\\
				$f_v^{S}$ & $f_t^{S_6}$ &\textbf{62.7} &\textbf{88.2} &\textbf{94.5} &\textbf{42.6} &\textbf{69.6} &\textbf{79.4} &\textbf{57.0} &\textbf{82.1}  &\textbf{88.8}\\
		\bottomrule
	\end{tabular}
	\caption{Comparisons of text-image retrieval results on MSCOCO~(1K and 5K) and Flickr 30K.}
	\label{table:main}
\end{table*}

\begin{table}[t!]
	\centering
	\begin{tabular}{c|ccc|c}
		\toprule
		\multirow{2}{*}{Image} &\multicolumn{3}{c|}{MSCOCO~(5K)} & $\Delta$\\
		&R@1 &R@5  &R@10  & R@1\\
		\cline{1-5}
		$f_v^{S}$  &\textbf{36.7} &\textbf{64.6} &\textbf{75.3} & -\\
		\cline{1-5}
		w/o stage-1 &32.6 &59.6 &70.7 &-4.1~ \\
		stage-1$_{\text{MSE}}$ &22.6  &46.7 &58.5 &-14.1 \\
		stage-1$_{\text{InfoNCE}}$ &31.7 &58.5 &69.6 &-5.0 \\
		\cline{1-5}
		w/o SF &30.9 &57.6 &70.8  &-5.8\\
		w/o KD &35.8 &63.1 &74.2 &-0.9\\
		w/o HN &34.4 &62.0 &73.7 &-2.3\\
		w/o KD+HN &32.6 &60.3 &71.9 &-4.1 \\
		\bottomrule
	\end{tabular}
	\caption{Ablation on design choices in both stages.}
	\label{table:main3}
\end{table}

\begin{table}[t!]
	\centering
	\begin{tabular}{cc|ccc}
		\toprule
		\multirow{2}{*}{Image}&\multirow{2}{*}{Text} & Disk Space &QPS$_v$ &QPS$_t$  \\
		&  &MB &\#  &\#  \\
		\cline{1-5}
		$f_v^{T^\prime}$ & $f_t^{T\prime}$ &578  &1.00x &1.00x \\
		\cline{1-5}
		$f_v^{S}$ & $f_t^{S_6}$ &255 &\textbf{1.51}x &1.98x \\
		$f_v^{S}$ & $f_t^{S_4}$ &\textbf{230} &\textbf{1.51}x &\textbf{2.77}x \\
		\bottomrule
	\end{tabular}
	\caption{Comparisons of disk space and QPS.}
	\label{table:main2}
\end{table}
\subsection{Setup}
\textbf{Dataset. }We use Conceptual Caption~\cite{cc} for stage-1 compression. 
It consists of 3M noisy image alt-text pairs. 
However, we do not use the image-text alignment information 
but only treat it as a reservoir of general-domain images and texts. In stage-2, we use MSCOCO~\cite{coco} and Flickr30K~\cite{flickr} as the benchmarks. For MSCOCO, there are 113,287 images for training, 5,000 images for validation, and both 5K and 1K for testing. For Flickr30K, there are 28,783 images for training, 1,000 images for validation, and 1k for testing.

\noindent
\textbf{Evaluation Metrics. }Following previous work, we use recall R@K~(K=1,5,10) as the main metric of task performance. We also report the disk space~(MB) and how many image/text queries can be encoded per second~(QPS$_v$ for image and QPS$_t$ for text) to evaluate model's memory footprints and inference speed.

\noindent
\textbf{Target Model.} We use the open-sourced ViT-B/32 CLIP as the target dual-encoder model to compress. The image encoder $f_v^{T}$ is a 12-layer Vision Transformer~\cite{vit} with 768 hidden dimension and 12 attention heads. The text encoder $f_t^{T}$ is a 12-layer Transformer with 512 hidden dimention and 8 attention heads. Note that this is the largest publically available version according to OpenAI's official repository.

\noindent
\textbf{Compression Configuration.} For image encoder $f_v^{S}$, we use a ViT-S/16 with 384 hidden dimension. We initialize it with weights pretrained on ImageNet-21K~\cite{imagenet21k}for faster convergence and better performance. For text encoder $f_t^{S}$, we experiment with both 6-layer and 4-layer Transformer~(marked as $f_t^{S_6}$ and $f_t^{S_4}$), of which the weights are initialized from corresponding layers in $f_t^{T}$. We also compare with 
a baseline compression method that directly fine-tunes pre-trained ViT-S/16 and 4-layer TinyBERT~\cite{tinybert} $f_t^{\text{TinyBERT}}$ using InfoNCE loss throughout both stages.

\textbf{Implementation Detail.} In stage-1, we train 1 epoch using AdamW~\cite{adamw} with a batch size of 84 for both images and texts, learning rate of 3e-4, and weight decay of 0.1. In stage-2, we use the same optimization setting except that we train with batch size 96 for 5 epochs. We employ a cosine learning rate scheduler with 10,000 warm-up steps for both stages. All reported results are calculated on the test set using checkpoints with the highest validation performance.

\subsection{Results}

\textbf{Main Results.} \tabref{table:main} summarizes the main results. As can be observed, the pre-trained CLIP model can already deliver moderately good retrieval performance. The performance is further improved after fine-tuning. Fine-tuning pre-trained ViT-S/16 and TinyBERT underperforms the zero-shot CLIP, showing that training with \textit{inter-modal} InfoNCE is not effective without extremely large-scale paired data. On most evaluation metrics, models compressed by our proposed two-stage pipeline perform on par with or better than the fine-tuned target model. We also found that the capacity of text encoder has limited affect on the performance.

\textbf{Ablation Study.} We perform extensive ablations to study the importance of each proposed technique. Due to the computational budget, we only conduct ablation on the image encoder and fix the text encoder as $f_t^{T^\prime}$. We evaluate  \textit{w/o stage-1}, \textit{stage-1$_{MSE}$}~(mean-square-error between $f_v^{S}$ and $f_v^{T}$), and \textit{stage-1$_{InfoNCE}$}~(identical to the loss in \secref{pre}) for stage-1 ablation. We also study the effectiveness of KD/SF/HN by removing them separately or together. We made several observations based on \tabref{table:main3}: 1) SF makes fine-tuning stable and is essential for convergence. 2) both KD and HN improve retrieval accuracy and are complementary to each other. 3) intra-modal contrastive distillation helps when image-text pairs are noisy and outperforms inter-modal infoNCE loss.

\textbf{Efficiency.} In \tabref{table:main2}, we compare the disk space and QSP used by models on a RTX 2080Ti of 12GB memroy. The compressed image encoder $f_v^{S}$ takes 85MB disk space~(39\% of $f_v^{T}$) meanwhile being 1.51x times faster. Our compressed text encoder can achieve up to x2.77 inference speed-up and 40\% size reduction~(from 243MB to 146MB). We further benchmark models'  memory and run-time performance on a real iPhone X with 1,000 images in the gallery for testing. It takes 870MB and 295MB for loading CLIP and our compressed model into main memory respectively. After indexing, the response time for a single text query is 0.4s for CLIP while it is only 0.1s for our compressed model. Although the results are hardward-dependent, our compressed model still shows an evident improvement in efficiency.


\section{Conclusion}
In this paper, we present a two-stage framework for lightweight text-image retrieval. Experiments on two benchmarks show the effectiveness of each component in the framework and the best performance is achieved when combining them together. It holds the merit of reducing model size and accelerating inference time, making memory/response-sensitive applications more practical.
\section*{Acknowledgement}
This research is partially supported by NSFC Grant No. 91646205, and SJTU-CMBCC Joint Research Scheme.
\bibliography{acl2022}
\bibliographystyle{acl_natbib}

\end{document}